%% file: 00-main.tex
\documentclass[11pt,a4paper]{article}
\usepackage[hyperref]{acl2017}
\usepackage{times}
\usepackage{latexsym}

\usepackage{url}

\usepackage{todonotes}
\usepackage{multirow}
\usepackage{verbatim}
\usepackage{graphicx}
\usepackage{amsfonts}  
\usepackage{bm}
\usepackage{subcaption}
\usepackage{amsmath}
\usepackage[belowskip=-5pt,aboveskip=0pt]{caption}

\aclfinalcopy 
\def\aclpaperid{3474} 

\title{
Question Answering through Transfer Learning \\ from Large Fine-grained Supervision Data
}

\author{Sewon Min\thanks{\hspace{2.5mm}All work was done while the author was an exchange student at University of Washington.} \\
  Seoul National University\\
  {\tt shmsw25@snu.ac.kr} \\\And
  Minjoon Seo \\
  University of Washington \\
  {\tt minjoon@uw.edu} \\\And
  Hannaneh Hajishirzi \\
  University of Washington \\
  {\tt hannaneh@uw.edu} \\}

\date{}

\begin{document}
\maketitle
\begin{abstract}
  We show that the task of question answering (QA) can significantly benefit from the transfer learning of models trained on a different large, fine-grained QA dataset.
  We achieve the state of the art in two well-studied QA datasets, WikiQA and SemEval-2016 (Task 3A), through a basic transfer learning technique from SQuAD.
  For WikiQA, our model outperforms the previous best model by more than 8\%.
  We demonstrate that finer supervision provides better guidance for learning lexical and syntactic information than coarser supervision, through quantitative results and visual analysis.
  We also show that a similar transfer learning procedure  achieves  the state of the art on an entailment task.

\end{abstract}

\section{Introduction}\label{sec:intro}
\input{01-intro}
\section{Background and Data}\label{sec:background}
\input{02-related}

\section{Model}\label{sec:model}
\input{04-model}

\section{Experiments}\label{sec:exp}

\input{05-exp}

\section{Conclusion}\label{sec:conclusion}
In this paper, we show state-of-the-art results on WikiQA and SemEval-2016 (Task 3A) as well as an entailment task, SICK, outperforming previous results by 8\%, 1\%, and 2\%, respectively. We show that question answering with sentence-level supervision can greatly benefit from standard transfer learning of a question answering model trained on a large, span-level supervision. We additionally show that  such transfer learning can be applicable in other NLP tasks such as textual entailment. 

\section*{Acknowledgments}
This research was supported by the NSF (IIS 1616112), NSF (III 1703166), Allen Institute for AI (66-9175), Allen Distinguished Investigator Award, and Google Research Faculty Award. We thank the anonymous reviewers for their helpful comments.


\bibliography{00-main}
\bibliographystyle{acl_natbib}

\newpage

\appendix

\section{Training details}\label{sec:app-A}
\input{10-app-A}

\section{More Analysis}\label{sec:app-B}
\input{11-app-B}

\section{More Results}\label{sec:app-C}
\input{12-app-C}

\end{document}

%% file: 01-intro.tex
Question answering (QA) is a long-standing challenge in NLP,  and the community has introduced several paradigms and datasets for the task over the past few years. These paradigms differ from each other in the type of questions and answers and the size of the training data, from a few hundreds to millions of examples.

We are particularly interested in the context-aware QA paradigm, where the answer to each question can be obtained by referring to its accompanying context (paragraph or a list of sentences).
Under this setting, the two most notable types of supervisions are coarse \emph{sentence-level} and fine-grained \emph{span-level}. 
In sentence-level QA, the task is to pick sentences that are most relevant to the question among a list of candidates~\cite{yang2015wikiqa}.
In span-level QA, the task is to locate the smallest span in the given paragraph that answers the question~\cite{rajpurkar2016squad}.

In this paper, we address coarser, sentence-level QA through a standard transfer learning\footnote{
The borderline between transfer learning and domain adaptation is often ambiguous~\cite{mou2016transferable}.
We choose the term ``transfer learning'' because we also adapt the pretrained QA model to an entirely different task, RTE.  
}
technique of a model trained on a large, span-supervised QA dataset. 
%
We demonstrate that the target task not only benefits from the scale of the source dataset but also the capability of the fine-grained span supervision to better learn syntactic and lexical information. 

For the source dataset, we pretrain on SQuAD~\cite{rajpurkar2016squad}, a recently-released, span-supervised QA dataset.
For the source and target models, we adopt BiDAF~\cite{bidaf}, one of the top-performing models in the dataset's leaderboard.
For the target datasets, we evaluate on two  recent QA datasets, WikiQA~\cite{yang2015wikiqa} and SemEval 2016 (Task 3A)~\cite{alessandromoschitti2016semeval}, which possess sufficiently different characteristics from that of SQuAD. Our results show 8\% improvement in WikiQA and 1\% improevement in SemEval. 
In addition, we  report state-of-the-art results on recognizing textual entailment (RTE) in SICK~\cite{marelli2014sick}  with a similar transfer learning procedure. 

%% file: 02-related.tex
\begin{table*}[ht]
\begin{center}
\resizebox{2.1\columnwidth}{!}{
\begin{tabular}{|l|l||l|l|l||l|l|} 
 \hline
 \multicolumn{2}{|c||}{Span-level QA} & \multicolumn{3}{|c||}{Sentence-level QA} &
 \multicolumn{2}{|c|}{RTE}\\
 \hline
 &\multicolumn{1}{|c||}{\textit{SQuAD}} & &\multicolumn{1}{|c|}{\textit{WikiQA}} & \multicolumn{1}{|c||}{\textit{SemEval-2016 Task 3A}} & & \multicolumn{1}{|c|}{\textit{SICK}} \\
  \hline
\multirow{3}{*}{\textbf{Q}} &Which company made& \multirow{3}{*}{\textbf{Q}}
&&I saw an ad, data entry jobs online. It req-&&
Four kids are \\
&Spectre? & & Who made airbus &
uired we give a fee and they promise fixed& \textbf{P} &
doing backbends\\
&&&&amount every month. Is this a scam? &&
in the park.\\
  \hline
\multirow{5}{*}{\textbf{C}} &
Spectre (2015) is the&\multirow{2}{*}{\textbf{C1}}&
Airbus SAS is an aircraft manufacturing subsi-
& well  probably is so i be more careful if i&&
Four girls are\\
&24th James Bond film&&
diary of EADS, a European aerospace company.
&were u. Why you looking for online jobs & &
doing backbends\\
  \cline{3-5}
&produced by Eon&  \textbf{C2} &
Airbus began as an union of aircraft companies.
& SCAM!!!!!!!!!!!!!!!!!!!!!! & \textbf{H} &
and playing\\
  \cline{3-5}
&
Productions. It
&\multirow{2}{*}{\textbf{C3}}
&
Aerospace companies allowed the establishment
& Bcoz i got a baby and iam nt intrested to&&
outdoors.\\ 
&
features  (...)
&&
of a joint-stock company, owned by EADS.
&sent him in a day care. thats y iam (...)&& \\
\hline
\textbf{A} & ``Eon Productions'' & \textbf{A} & C1(Yes), C2(No), C3(No) & C1(Good), C2(Good), C3(Bad) & \textbf{A} & Entailment \\
\hline
\end{tabular}
}
\end{center}
\caption{\small Examples of question-context pairs from QA datasets and premise-hypothesis pair from RTE dataset. \textbf{Q} indicates question, \textbf{C} indicates context, \textbf{A} indicates answer, \textbf{P} indicates premise and \textbf{H} indicates hypothesis. } 
\label{tab:data}
\end{table*}

Modern machine learning models, especially deep neural networks, often significantly benefit from transfer learning.
In computer vision, deep convolutional neural networks trained on a large image classification dataset such as ImageNet \cite{deng2009imagenet} have proved to be useful for initializing models on other vision tasks, such as object detection~\cite{zeiler2014visualizing}.
In natural language processing, domain adaptation has traditionally been an important topic for syntactic parsing \cite{mcclosky2010automatic} and named entity recognition \cite{chiticariu2010domain}, among others.
With the popularity of distributed representation, pre-trained word embedding models such as word2vec \cite{mikolov2013distributed, mikolov2013efficient} and glove \cite{pennington2014glove} are also widely used for natural language tasks~\cite{karpathy2015deep,kumar2015ask}. Instead of these, we initialize our models from a QA dataset and show how standard transfer learning can achieve state-of-the-art in target QA datasets. 

There have been several QA paradigms in NLP, which can be categorized by the context and supervision used to answer questions. This context can range from structured and confined knowledge bases~\cite{berant2013semantic} to unstructured and unbounded natural language form (e.g.,  documents on the web~\cite{qasent}) and unstructured, but restricted in size (e.g., a paragraph or multiple sentences~\cite{hermann2015teaching}). The recent advances in neural question answering lead to numerous datasets and successful models in these paradigms~\cite{rajpurkar2016squad, yang2015wikiqa,nguyen2016ms,trischler2016newsqa}. The answer types in these datasets are largely divided into three categories: sentence-level, in-context span, and generation. In this paper, we specifically focus on the former two and show that span-supervised models can better learn syntactic and lexical features.
Among these datasets, we briefly describe three QA datasets to be used for the experiments in this paper.
We also give the description of an RTE dataset for an example of a non-QA task.
Refer to Table~\ref{tab:data} to see the examples of the datasets.





\vspace{-.1cm}
\paragraph{SQuAD}\cite{rajpurkar2016squad} is a recent span-based QA dataset, containing 100k/10k train/dev examples. Each example is a pair of context paragraph  from Wikipedia and a question created by a human, and the answer is a span in the context.

\vspace{-.1cm}
\paragraph{SQUAD-T} is our modification of SQuAD dataset to allow for sentence selection QA. (`T' for sen{\it T}ence).
We split the context paragraph into sentences and  formulate the task as classifying whether each sentence contains the answer. 
This enables us to make a fair comparison between pretraining with span-supervised and sentence-supervised QA datasets.

\vspace{-.1cm}
\paragraph{WikiQA}\cite{yang2015wikiqa} is a sentence-level QA dataset, containing 1.9k/0.3k train/dev answerable examples.
Each example consists of a real user's Bing query and a snippet of a Wikipedia article retrieved by Bing, containing 18.6 sentences on average. The task is to classify whether each sentence provides the answer to the query.

\vspace{-.1cm}
\paragraph{SemEval 2016 (Task 3A) }\cite{alessandromoschitti2016semeval} is a sentence-level QA dataset, containing 1.8k/0.2k/0.3k train/dev/test examples.
Each example consists of a community question by a user and 10 comments. The task is to classify whether each comment is relevant to the question.

\vspace{-.1cm}
\paragraph{SICK}\cite{marelli2014sick} is a dataset for recognizing textual entailment (RTE), containing 4.5K/0.5K/5.0K train/dev/test examples.
Each example consists of a hypothesis and a premise, and the goal is to determine if the premise is entailed by, contradicts, or is neutral to the hypothesis (hence classification problem).
We also report results on SICK to show that span-supervised QA dataset can be also useful for non-QA datasets. 


%% file: 04-model.tex
Among numerous models proposed for span-level QA tasks~\citep{xiong2016dynamic, wang2016machine},
we adopt an open-sourced model, BiDAF\footnote{\url{https://allenai.github.io/bi-att-flow}}~\citep{bidaf}.

\vspace{-.1cm}
\paragraph{BiDAF.} 
The inputs to the model are a question ${\bm q}$, and a context paragraph ${\bm x}$. Then the model selects the best answer span, which is $\arg \max_{(i,j)} {\bf y}^\mathrm{start}_i {\bf y}^\mathrm{end}_j$, where $i <= j$. Here, ${\bf y}^\mathrm{start}_i$ and ${\bf y}^\mathrm{end}_i$ are start and end position probabilities of $i$-th element, respectively.


Here, we briefly describe the answer module which is important for transfer learning to sentence-level QA. The input to the answer module is a sequence of vectors $\{{\bf h}_i\}$ each of which encodes enough information about the $i$-th context word and its relationship with its surrounding words and the question words.
Then the role of the answer module is to map each vector ${\bf h}_i$ to its start and end position probabilities, ${\bf y}^\mathrm{start}_i$ and ${\bf y}^\mathrm{end}_i$.

\vspace{-.1cm}
\paragraph{BiDAF-T} refers to the modified version of BiDAF to make it compatible with sentence-level QA\footnote{Code available at: \url{https://github.com/shmsw25/qa-transfer}}. (`T' for sen{\it T}ence).
In this task, the inputs are a question ${\bm q}$ and a list of sentences, ${\bm x}_1, \dots, {\bm x}_T$, where $T$ is the number of the sentences.
Note that, unlike BiDAF, which outputs single answer per example, 
Here we need to output a $C$-way classification for each $k$-th sentence. 

Since BiDAF is a span-selection model, it cannot be directly used for sentence-level classification.
Hence we replace the original answer module of BiDAF with a different answer module, and keep the other modules identical to those of BiDAF.
Given the input to the new answer module, $\{ {\bf h}^k_1, \dots, {\bf h}^k_N\}$, where the superscript is the sentence index ($1 \leq k \leq T$), we obtain the C--way classification scores for the $k$-th sentence, ${\tilde {\bf y}}^k \in [0,1]^C$ via max-pooling method:
\begin{equation}
{\tilde {\bf y}}^k = \mathrm{softmax} ({\bf W} \max({\bf h}^k_1, \dots, {\bf h}^k_N) + {\bf b})
\end{equation}
where ${\bf W} \in \mathbb{R}^{C \times d}, {\bf b} \in \mathbb{R}^C$ are trainable weight matrix and bias, respectively, and $\max()$ function is applied elementwise.

For WikiQA and SemEval 2016, the number of classes ($C$) is $2$, i.e. each sentence (or comment) is either relevant or not relevant.
Since some of the metrics used for these datasets require full ranking, we use the predicted probability for ``relevant'' label to rank the sentences.

Note that BiDAF-T can be also used for the RTE dataset, where we can consider the hypothesis as a question and the premise as a context sentence ($T=1$), and classify each example into `entailment', `neutral', or `contradiction' ($C=3$).


\vspace{-.1cm}
\paragraph{Transfer Learning.}
Transfer learning between the same model architectures\footnote{
Strictly speaking, this is a domain adaptation scenario.
} is straightforward: we first initialize the weights of the target model with the weights of the source model pretrained on the source dataset, and  then we further train (finetune) on the target model with the target dataset. 
To transfer from BiDAF (on SQuAD) to BiDAF-T, we transfer all the weights of the identical modules, and initialize the new answer module in BiDAF-T with random values. 
For more training details, refer to Appendix~\ref{sec:app-A}.

%% file: 05-exp.tex
\vspace{-.1cm}
\begin{table}[ht]
\begin{center}
\resizebox{\columnwidth}{!}{
\begin{tabular}{|c|c||c|c|c||c|c|c|}
\hline
Pretrained & Fine- & \multicolumn{3}{|c||}{WikiQA} & \multicolumn{3}{|c|}{SemEval-2016} \\
\cline{3-8}
dataset & tuned& MAP & MRR & P@1 & MAP & MRR & AvgR\\
\hline
- & - & 62.96  & 64.47 & 49.38 & 76.40  & 82.20 & 86.51\\
SQuAD-T & No & 75.22 & 76.40 & 62.96 & 47.23 & 49.31 & 60.01 \\
SQuAD & No & 75.19 & 76.31 & 62.55 & 57.80 & 66.10 & 71.13 \\
SQuAD-T & Yes & 76.44 & 77.85 & 64.61 & 76.30 & 82.51 & 86.64 \\
SQuAD & Yes & 79.90 & 82.01 & 70.37 & 78.37 & 85.58 & 87.68 \\
SQuAD* & Yes & {\bf83.20} & {\bf84.58}& {\bf75.31}  & {\bf80.20} & {\bf86.44} &{\bf89.14} \\
\hline
\multicolumn{2}{|c||}{Rank 1} & 74.33 & 75,45 & -& 79.19 & 86.42 &  88.82 \\
\multicolumn{2}{|c||}{Rank 2} & 74.17 & 75.88 & 64.61 & 77.66 & 84.93 &  88.05 \\
\multicolumn{2}{|c||}{Rank 3} & 70.69 & 72.65  & - & 77.58 & 85.21 &  88.14 \\
\hline
\end{tabular}
}
\end{center}
\caption{ \small
Results on WikiQA and SemEval-2016 (Task 3A). The first row is a result from non-pretrained model, and * indicates ensemble method.
Metrics used are Mean Average Precision (MAP), Mean Reciprocal Rank (MRR), Precision at rank 1 (P@1), and Average Recall (AvgR).
Rank 1,2,3 indicate the results by previous works, ordered by MAP. For WikiQA, they are from \citet{wang2016compare,Tymoshenko2016Convolutional,miller2016key}, respectively. For SemEval-2016, they are from \citet{filice2016kelp,joty2016convkn, mihaylov2016semanticz}.
SQuAD*\&Yes sets the new state of the art on both datasets.
}
\label{tab:result}
\end{table}

\vspace{-.1cm}
\paragraph{Question Answering Results.}
\label{subsec:qa}

Table~\ref{tab:result} reports the state-of-the-art results of our transfer learning on WikiQA and SemEval-2016 and the performance of previous models as well as several ablations that use no pretraining or no finetuning.
There are multiple interesting observations from Table~\ref{tab:result} as follows:

\vspace{-.2cm}
\paragraph{(a)} If we only train the BiDAF-T model on the target datasets with no pretraining (first row of Table~\ref{tab:result}), the results are poor. This shows the importance of both pretraining and finetuning.

\vspace{-.2cm}
\paragraph{(b)} Pretraining on SQuAD and SQuAD-T with no finetuning (second and third row) achieves results close to the state-of-the-art in the WikiQA dataset, but not in SemEval-2016. Interestingly, our result on SemEval-2016 is not better than only training without transfer learning. We conjecture that this is due to the significant difference between the domain of SemEval-2016 and that of SQuAD, which are from community and Wikipedia, respectively.

\vspace{-.2cm}
\paragraph{(c)} Pretraining on SQuAD and SQuAD-T with finetuning (fourth and fifth row) significantly outperforms (by more than 5\%) the highest-rank systems on WikiQA. It also outperforms the second ranking system in SemEval-2016 and is only 1\% behind the first ranking system.

\vspace{-.2cm}
\paragraph{(d)} Transfer learning models achieve better results with pretraining on span-level supervision (SQuAD) than coarser sentence-level supervision (SQuAD-T).\footnote{We additionally perform Mann-Whitney U Test and McNemar’s Test to show the statistical significance of the advantage of span-level pretraining over sentence-level pretraining. For WikiQA, the advantage is statistically significant with the confidence levels of 97.1\% and 99.6\%, respectively. For SemEval, we obtain the confidence levels of 97.8\% and 99.9\%, respectively.}

\vspace{+.2cm}

Finally, we also use the ensemble of 12 different training runs on the same BiDAF architecture, which obtains the state of the art in both datasets. This system outperforms the highest-ranking system in WikiQA by more than 8\% and the best system in SemEval-2016 by 1\% in every metric. It is important to note that, while we definitely benefit from the scale of SQuAD for transfer learning to smaller WikiQA, given the gap between SQuAD-T and SQuAD ($>3\%$), we see a clear sign that span-supervision plays a significant role well.

\vspace{-.1cm}
\paragraph{Varying the size of pretraining dataset.}
\label{subsec:ablations}

We vary the size of SQuAD dataset used during pretraining, and test on WikiQA with finetuning. Results are shown in Table~\ref{tab:varying-size}. As expected, MAP on WikiQA drops as the size of SQuAD decreases. 
It is worth noting that pretraining on SQuAD-T (Table~\ref{tab:result}) yields 0.5 point lower MAP than pretraining on 50\% of SQuAD.
In other words, roughly speaking, span-level supervision data is worth more than twice the size of sentence-level supervision data for the purpose of pretraining.
Also, even a small size of fine-grained supervision data helps; pretraining with 12.5\% of SQuAD gives an advantage of more than 7 points than no pretraining.

\begin{table}[t]
\begin{center}
\resizebox{0.8\columnwidth}{!}{
\begin{tabular}{cc}
\hline
Percentage of used SQuAD dataset & MAP \\
\hline
\hline
100\% & 79.90 \\
50\% & 76.94 \\
25\% & 74.39 \\
12.5\% & 70.76  \\
\hline
\end{tabular}
}
\end{center}
\caption{ \small 
Results with varying sizes of SQuAD dataset used during pretraining. All of them are finetuned and tested on WikiQA.
}
\label{tab:varying-size}
\end{table}

\begin{figure}[t]
\centering
\resizebox{\columnwidth}{!}{
\includegraphics[width=\textwidth]{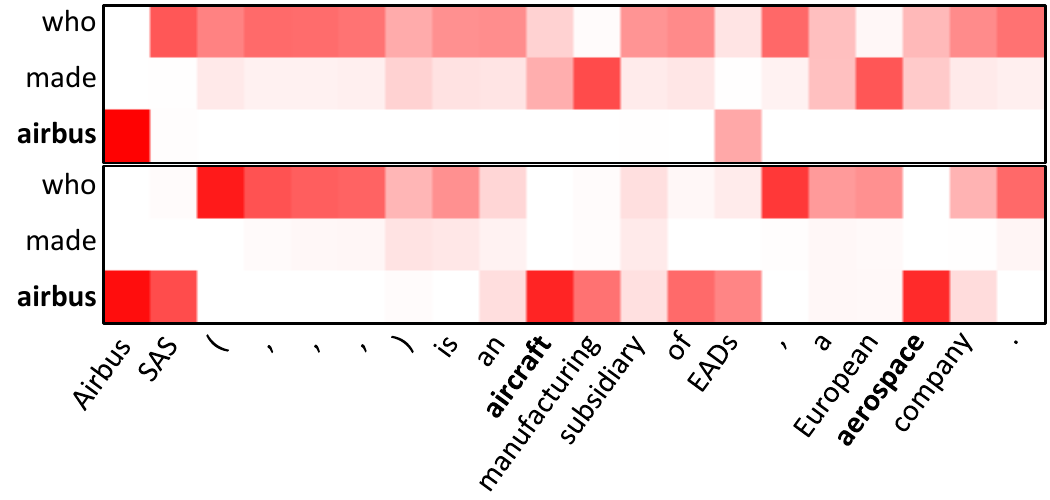}
}
\caption{\small Attention maps showing correspondence between the words of a question (vertical) and the subset of its context (horizontal) in WikiQA for (top) SQuAD-T-pretrained model and (bottom) SQuAD-pretrained model. 
The more red, the higher the correspondence.
}
\label{fig:wikiqa-attmap}
\end{figure}

\begin{figure}[ht]
\centering
\resizebox{\columnwidth}{!}{
\includegraphics[width=0.6\textwidth]{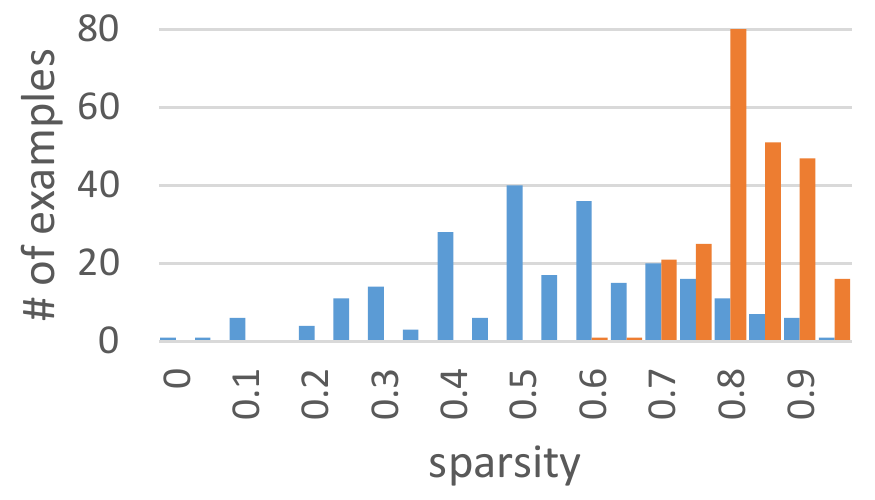}
}
\caption{\small Histogram of the sparsity (Equation~\ref{eqn:sparsity}) of the attention maps of SQuAD-T-pretrained model (SQuAD-T\&Yes, blue) and SQuAD-pretrained model (SQuAD\&Yes, red).
Mean sparsity of SQuAD-pretrained model ($0.84$) is clearly higher than that of SQuAD-T-pretrained model ($0.56$).
}
\label{fig:sparsity}
\end{figure}

\vspace{-.1cm}
\paragraph{Analysis.}
\label{subsec:varying-size}

Figure~\ref{fig:wikiqa-attmap} shows the latently-learned attention maps between the question and one of the context sentences from a WikiQA example in Table~\ref{tab:data}.
The top map is pretrained on SQuAD-T (corresponding to SQuAD-T\&Yes in Table~\ref{tab:result}) and the bottom map is pretrained on SQuAD (SQuAD\&Yes).
The more red the color, the higher the relevance between the words.
There are two interesting observations here.

First, in SQuAD-pretrained model (bottom), we see a high correspondence between question's \texttt{airbus} and context's \texttt{aircraft} and  \texttt{aerospace}, but the SQuAD-T-pretrained model  fails to learn such correspondence.

Second, we see that the attention map of the SQuAD-pretrained model is more sparse, indicating that it is able to more precisely localize correspondence between question and context words.
In fact, we compare the sparsity of WikiQA test examples in SQuAD\&Y and SQuAD-T\&Y. Following~\citet{hurley2009comparing}, the sparsity of an attention map is defined by
\begin{equation}\label{eqn:sparsity}
\mathrm{sparsity} = \dfrac{|\left\{ x\in\mathbb{V}|x\leq \epsilon \right\}|}{|\mathbb{V}|}
\end{equation}
where $\mathbb{V}$ is a set of values between $0$ and $1$ in attention map, and $\epsilon$ is a small value which we define $0.01$ for here.
A histogram of the sparsity is shown in Figure~\ref{fig:sparsity}. There is a large gap in the average sparsity of WikiQA test examples between SQuAD\&Yes and SQuAD-T\&Yes, which are $0.84$ and $0.56$, respectively.

More analyses including error analysis and more visualizations are shown in Appendix~\ref{sec:app-B}.




\vspace{-.1cm}
\paragraph{Entailment Results.}
\label{subsec:rte}

\begin{table}[t]
\begin{center}
\resizebox{0.8\columnwidth}{!}{
\begin{tabular}{cc}
\hline
Pretrained dataset / Previous work & Accuracy \\
\hline
\hline
- & 77.96 \\
SQuAD-T & 81.49 \\
SQuAD & 82.86 \\
SQuAD* & 84.38  \\
\hline
SNLI &  83.20 \\
SQuAD-T $+$ SNLI & 85.00 \\
SQuAD $+$ SNLI & 86.63 \\
SQuAD $+$ SNLI* & {\bf 88.22} \\
\hline
\citet{yin2015abcnn} &  86.2 \\
\citet{lai2014illinois} & 84.57  \\
\citet{zhao2014ecnu} & 83.64 \\
\citet{jimenez2014unal} & 83.05  \\
\hline
\citet{mou2016transferable} & 70.9  \\
\citet{mou2016transferable} (pretrained on SNLI) & 77.6  \\
\hline
\end{tabular}
}
\end{center}
\caption{ \small 
Results on SICK after finetuning. The first row is only trained on SICK.  * indicates ensemble method.
}
\label{tab:result-sick}
\end{table}
In addition to QA experiments, we also show that the models trained on span-supervised QA can be useful for textual entailment task (RTE). 
Table~\ref{tab:result-sick} shows the transfer learning results of BiDAF-T on SICK dataset \citep{marelli2014sick}, with various pretraining routines.
Note that SNLI~\cite{bowman2015large} is a similar task to SICK and is significantly larger (150K/10K/10K train/dev/test examples).
Here we highlight three observations:

\vspace{-.2cm}
\paragraph{(a)} BiDAF-T pretrained on SQuAD outperforms that without any pretraining by 6\% and that pretrained on SQuAD-T by 2\%, which demonstrates that the transfer learning from large span-based QA gives a clear improvement.

\vspace{-.2cm}
\paragraph{(b)} Pretraining on SQuAD+SNLI outperforms pretraining on SNLI only.
Given that SNLI is larger than SQuAD, the difference in their performance is a strong indicator that we are benefiting from not only the scale of SQuAD, but also the fine-grained supervision that it provides.

\vspace{-.2cm}
\paragraph{(c)} We outperform the previous state of the art by 2\% with the ensemble of SQuAD+SNLI pretraining routine.

It is worth noting that~\citet{mou2016transferable} also shows improvement on SICK by pretraining on SNLI.

%% file: 10-app-A.tex
\vspace{-.1cm}
\paragraph{Parameters.}
For pretraining BiDAF on SQuAD, we  follow the exact same procedure in \citet{bidaf}. For pretraining BiDAF-T on SQuAD-T, we use the same hyperparameters for all modules except the answer module, for which we use the hidden state size of $200$. The learning rate is controlled by AdaDelta~\cite{zeiler2012adadelta} with the initial learning rate of $0.5$ and minibatch size of $50$. 
We maintain the moving averages of all weights of the model with the exponential decay rate of 0.999 during training and use them at test. The loss function is the cross entropy between $\tilde {\bf y}^k$ and the one-hot vector of the correct classification.

\vspace{-.1cm}
\paragraph{Convergence.}
For all settings, we train models until performance on development set continue to decrease for 5k steps.
Table~\ref{tab:convergence} shows the median selected step on each setting.

\begin{table}[ht]
\begin{center}
\resizebox{0.8\columnwidth}{!}{
\begin{tabular}{|c|c||c|}
\hline
Dataset & Pretrained & selected step \\
\hline
SQuAD & - & 18k \\
SQuAD-T & - & 50k \\
\hline
WikiQA & - & 6k \\
WikiQA & SQuAD-T & 6k \\
WikiQA & SQuAD & 3k \\
\hline
SemEval-2016 & - & 9k \\
SemEval-2016 & SQuAD-T & 4k \\
SemEval-2016 & SQuAD & 3k \\
\hline
SICK & - & 13k \\
SNLI & - & 55k \\
SICK & SQuAD-T & 9k \\
SNLI & SQuAD-T & 31k \\
SICK & SQuAD & 18k \\
SNLI & SQuAD & 49k \\
SICK & SQuAD-T $+$ SNLI & 7k \\
SICK & SQuAD $+$ SNLI & 7k \\
\hline
\end{tabular}
}
\end{center}
\caption{ \small Median global step, which has the best performance on development set.
}
\label{tab:convergence}
\end{table}

%% file: 11-app-B.tex
\begin{table*}
\begin{center}
\begin{tabular}{c} 

\begin{subtable}{\textwidth}
\resizebox{\columnwidth}{!}{
\begin{tabular}{|c|c|l|l|} 
 \hline
 Category Id & Category & Example Question & Relevant Sentence \\ 
\hline \hline
\multirow{2}{*}{1}   & \multirow{2}{*}{Exact Match}  & \multirow{2}{*}{When did SpongeBob first air?}  &  The pilot episode of SpongeBob SquarePants first aired in the  \\
&&& United States on May 1, 1999, following the ...  \\
 \hline
2   & Paraphrase  & When was How the West Was Won filmed?  & How the West Was Won is a 1962 American epic – Western film. \\  \hline
3   & No Clear Clue  & When was Mary Anderson born? & Mary Anderson (1866-1953) was a real estate developer, ....\\
  \hline
4   & Need prior sentence (pronoun)  & When did Texas become a state?  & In 1845, it joined the United States as the 28th state. \\  \hline
\multirow{3}{*}{5}   & \multirow{3}{*}{Need prior sentence (context)} & \multirow{2}{*}{How do you play spades?}  & Its major difference as compared to other Whist variants is that, \\ &&& instead of trump being decided by the highest bidder or at random, \\
&&& the Spade suit is always trump, hence the name.  \\  \hline
6   & Hard to answer & How kimberlite pipes form?  &  Volcanic pipes are relatively rare. \\\hline
\end{tabular}
}
\end{subtable}
\\
\\
\begin{subtable}{\textwidth}
\resizebox{\columnwidth}{!}{
\begin{tabular}{|c|c|c|l|} 
 \hline
 Category Id & Category & Example Question Id & Example Question \\ 
\hline
  \hline
1   & Asking information & Q347\textunderscore R25 & hi all is there any IKEA showroom in and around DOHA? Kindly reply thank you \\  \hline
\multirow{3}{*}{2}  & Asking opinion or  & \multirow{3}{*}{Q326\textunderscore R90} & Salam  I am mechanical Eng. 15 years experience  i got a job for Rasgas co. direct \\
& recommendation in  & & hire I am married and i have 2 kids  5 and 3 years old.my life style \\
&specific situation&& is average. Is 8.000 QR enough as a basic salary? (...) \\
   \hline
\multirow{2}{*}{3}   & Asking feelings in & Q348\textunderscore R67 & oh i wish they will build Disneyland in Qatar : ) how do you think guys? It will be \\
&specific situation&& perfect for Qatar : )
 \\
   \hline
\multirow{2}{*}{4}   & \multirow{2}{*}{Asking abstract thing} & Q341\textunderscore R11 & i’d like to get to know more about Al Jazeera International from anyone on QATAR \\
&&& LIVING who works at Al Jazeera.
 \\
   \hline
\multirow{2}{*}{5}   & \multirow{2}{*}{Not Asking} & \multirow{2}{*}{Q337\textunderscore R21} & I just stumbled across this news article about the the American university campuses \\
&&& at Education City and thought some of you may also find it interesting.
 \\
   \hline
\multirow{3}{*}{6}   & \multirow{3}{*}{Asking a lot of things at once} & \multirow{2}{*}{Q337\textunderscore R16} & How good are Karwa services? Are they : 1. Courteous/Rude? 2. Taking the correct \\
&&&route/Longer route? (...) 7. A \\
&&& pleasure/displeasure to ride?
 \\
\hline
\end{tabular}
}
\end{subtable}
\end{tabular}
\end{center}
\caption{ \small Examples from each category on (top) WikiQA and (bottom) SemEval-2016 (Task 3A).}
\label{tab:category}
\end{table*}


\begin{figure}[!tb]
\centering
\resizebox{\columnwidth}{!}{
\includegraphics[width=\textwidth]{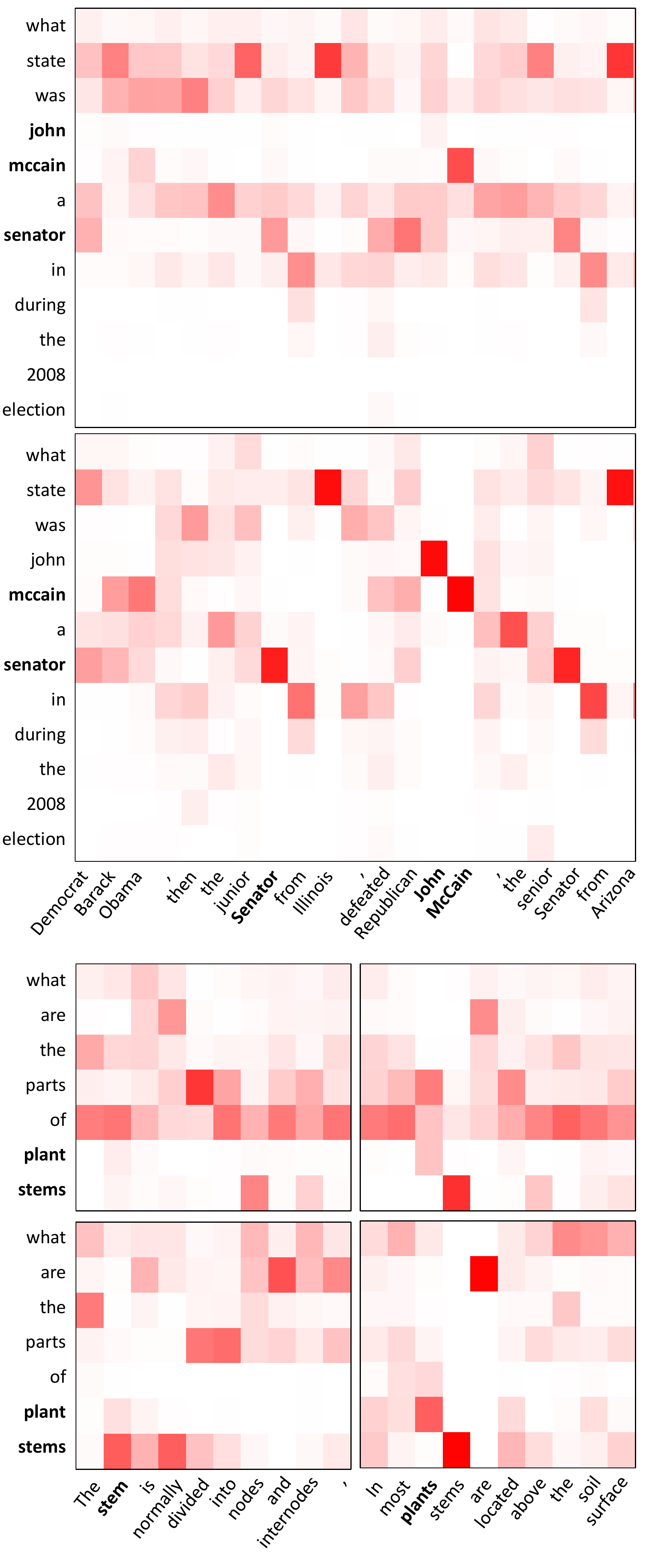}
}
\caption{\small
More attention maps showing correspondence between the words of a question (vertical) and one of candidate sentences (horizontal) in WikiQA for (top in each subfigure) SQuAD-MC-pretrained model and (bottom in each subfigure) SQuAD-pretrained model. 
The more red, the higher the correspondence.
}
\label{fig:att-more}
\end{figure}

\begin{table}[ht]
\begin{center}
\resizebox{\columnwidth}{!}{
\begin{tabular}{|c|c|c|}
\hline
 & WikiQA & SemEval-2016 \\
\hline
SQuAD-T\&Yes & C2 $>$ C1 $>$ C3 & C1 $>$ C3 $>$ C2 \\
SQuAD\&Yes & C1 $>$ C2 $>$ C3 & C2 $>$ C1 $>$ C3 \\
Groundtruth & C1(Y), C2(N), C3(N) & C1(Good), C2(Good), C3(Bad)\\
\hline
\end{tabular}
}
\end{center}
\caption{ \small Comparison of ranked answers by SQuAD-T-pretrained model (SQuAD-T\&Yes) and SQuAD-pretrained model (SQuAD\&Yes) of examples from WikiQA and SemEval-2016 (Task 3A) in Table~\ref{tab:data}.
}
\label{tab:error}
\end{table}

\begin{table}[!ht]
\begin{center}
\resizebox{\columnwidth}{!}{
\begin{tabular}{|c|c||c|c|c|c|c|c|c|}
\hline
\multicolumn{2}{|c||}{Pretrained dataset} & \multirow{2}{*}{total} & \multicolumn{6}{|c|}{Category Id}\\
\cline{1-2} \cline{4-9}
SQuAD-T-Y & SQuAD-Y &  &1 &2 &3 &4&5& 6 \\
\hline
\multicolumn{2}{|c||}{total} & 100 & 37 & 38 & 6 & 15 & 2 & 2\\
\hline
Correct & Correct & 49 & 28 & 14 & 3 & 4 & 0 & 0 \\
Wrong& Correct & \bf{26} & \bf{8} & \bf{14} & 1 & 3 & 0 & 0\\
Correct & Wrong & \bf{9} & \bf{1} & \bf{4} & 1 & 3  & 0 & 0\\
Wrong & Wrong & 16 & 0 & 6  & 1 & 5 & 2 & 2\\
\hline
\end{tabular}
}
\end{center}
\caption{ \small Comparison of performance of SQuAD-T-pretrained model (SQuAD-T-Y) and SQuAD-pretrained model (SQuAD-Y) on WikiQA.
}
\label{tab:wikiqa-comparison}
\end{table}

\begin{table}[!ht]
\begin{center}
\resizebox{\columnwidth}{!}{
\begin{tabular}{|c|c||c|c|c|c|c|c|c|}
\hline
\multicolumn{2}{|c||}{Pretrained dataset} & \multirow{2}{*}{total} & \multicolumn{6}{|c|}{Category Id}\\
\cline{1-2} \cline{4-9}
No Pretrain & SQuAD-Y & &1 &2 &3 &4 &5 & 6 \\
\hline
\multicolumn{2}{|c||}{total} & 100 & 29 & 38 & 7 & 12 & 9 & 5\\
\hline
Correct & Correct & 30 & 12 & 11 & 2 & 5 & 0 & 0 \\
Wrong & Correct & \bf{22} & \bf{6} & \bf{10} & 0 & 2 & 2 & 2\\
Correct & Wrong & \bf{5} & \bf{0} & \bf{1} & 2 & 1  & 0 & 1\\
Wrong & Wrong & 43 & 11 & 16  & 3 & 4 & 7 & 2\\
\hline
\end{tabular}
}
\end{center}
\caption{ \small Comparison of performance of model without pretraining (No Pretrain) and SQuAD-pretrained model (SQuAD-Y) on SemEval-2016 (Task 3A).
}
\label{tab:semeval-comparison}
\end{table}

\vspace{-.1cm}
\paragraph{Attention maps.}
We show some more examples of attention maps in Figure~\ref{fig:att-more}.
(Top) We see high correspondence between same word from question and context such as  \texttt{senator} and \texttt{john}, in SQuAD-pretrained model, but the SQuAD-T-pretrained model fails to learn such correspondence. (Bottom) We see high correspondence between \texttt{stems} from question and \texttt{stem} from context (left) as well as \texttt{plant} from question and \texttt{plants} from context (right), in SQuAD-pretrained model, but the SQuAD-T-pretrained model fails to learn such correspondence.

\vspace{-.1cm}
\paragraph{Error Analysis.}
Table~\ref{tab:error} shows the comparison between answers by SQuAD-T-pretrained model and SQuAD-pretrained model on the example of WikiQA and SemEval-2016 from Table~\ref{tab:data}. On WikiQA, SQuAD-T-pretrained model selects C2 instead of the groundtruth answer C1. On SemEval-2016, SQuAD-pretrained model ranks C3 (bad comment) higher than C2 (good comment).

In addition, we sampled 100 example randomly from WikiQA and SemEval-2016, and classified them into 6 categories(Table~\ref{tab:category}). 
In Table~\ref{tab:wikiqa-comparison}, we compare the performance on these WikiQA examples by SQuAD-T-pretrained model and SQuAD-pretrained model.
It shows that span supervision clearly helps answering questions on Category 1 and 2, which are easier to answer, with answering correctly on most of the questions in Category 1.
Similarly, we show the comparison of the performance on classified examples of the model without pretraining and SQuAD-pretrained model on SemEval-2016.
It also shows that span supervision helps answering questions asking information or opinion/recommendation.

%% file: 12-app-C.tex
\vspace{-.1cm}
\paragraph{SQuAD-T.}
To better understand SQuAD-T dataset,
we show the performance BiDAF-T with different training routines.
We get MAP 89.46 and accuracy 85.34\% with SQuAD-trained BiDAF model, and MAP 90.18 and accuracy 84.69\% with SQuAD-T-trained BiDAF-T model.
There is no large gap between the two models, as each paragraph of SQuAD-T has 5 sentences on average, which makes the classification problem easier than WikiQA.

\paragraph{SNLI.}Other larger RTE datasets such as SNLI also benefit from transfer learning, although the improvement is smaller.
We confirm the improvement by showing that the result on SNLI when pretraining on SQuAD with BiDAF is 82.6\%, which is slightly higher than that of the model pretrained on SQuAD-T (81.6\%).
This, however, did not outperform the state of the art (88.8\%) by \citet{wang2017bilateral}. This is mostly because BiDAF (or BiDAF-T) is a QA model, which is not designed for RTE tasks.

%% file: 00-main.bbl
\begin{thebibliography}{}
\expandafter\ifx\csname natexlab\endcsname\relax\def\natexlab#1{#1}\fi

\bibitem[{Berant et~al.(2013)Berant, Chou, Frostig, and
  Liang}]{berant2013semantic}
Jonathan Berant, Andrew Chou, Roy Frostig, and Percy Liang. 2013.
\newblock Semantic parsing on freebase from question-answer pairs.
\newblock In {\em EMNLP\/}.

\bibitem[{Bowman et~al.(2015)Bowman, Angeli, Potts, and
  Manning}]{bowman2015large}
Samuel~R Bowman, Gabor Angeli, Christopher Potts, and Christopher~D Manning.
  2015.
\newblock A large annotated corpus for learning natural language inference.
\newblock In {\em EMNLP\/}.

\bibitem[{Chiticariu et~al.(2010)Chiticariu, Krishnamurthy, Li, Reiss, and
  Vaithyanathan}]{chiticariu2010domain}
Laura Chiticariu, Rajasekar Krishnamurthy, Yunyao Li, Frederick Reiss, and
  Shivakumar Vaithyanathan. 2010.
\newblock Domain adaptation of rule-based annotators for named-entity
  recognition tasks.
\newblock In {\em EMNLP\/}.

\bibitem[{Deng et~al.(2009)Deng, Dong, Socher, Li, Li, and
  Fei-Fei}]{deng2009imagenet}
Jia Deng, Wei Dong, Richard Socher, Li-Jia Li, Kai Li, and Li~Fei-Fei. 2009.
\newblock Imagenet: A large-scale hierarchical image database.
\newblock In {\em CVPR\/}.

\bibitem[{Filice et~al.(2016)Filice, Croce, Moschitti, and
  Basili}]{filice2016kelp}
Simone Filice, Danilo Croce, Alessandro Moschitti, and Roberto Basili. 2016.
\newblock Kelp at semeval-2016 task 3: Learning semantic relations between
  questions and answers.
\newblock {\em SemEval\/} 16:1116--1123.

\bibitem[{Hermann et~al.(2015)Hermann, Kocisky, Grefenstette, Espeholt, Kay,
  Suleyman, and Blunsom}]{hermann2015teaching}
Karl~Moritz Hermann, Tomas Kocisky, Edward Grefenstette, Lasse Espeholt, Will
  Kay, Mustafa Suleyman, and Phil Blunsom. 2015.
\newblock Teaching machines to read and comprehend.
\newblock In {\em NIPS\/}.

\bibitem[{Hurley and Rickard(2009)}]{hurley2009comparing}
Niall Hurley and Scott Rickard. 2009.
\newblock Comparing measures of sparsity.
\newblock {\em IEEE Transactions on Information Theory\/} 55(10):4723--4741.

\bibitem[{Jimenez et~al.(2014)Jimenez, Duenas, Baquero, Gelbukh, B{\'a}tiz, and
  Mendiz{\'a}bal}]{jimenez2014unal}
Sergio Jimenez, George Duenas, Julia Baquero, Alexander Gelbukh, Av~Juan~Dios
  B{\'a}tiz, and Av~Mendiz{\'a}bal. 2014.
\newblock Unal-nlp: Combining soft cardinality features for semantic textual
  similarity, relatedness and entailment.
\newblock In {\em SemEval Workshop\/}.

\bibitem[{Joty et~al.(2016)Joty, Moschitti, Al~Obaidli, Romeo, Tymoshenko, and
  Uva}]{joty2016convkn}
Shafiq Joty, Alessandro Moschitti, Fahad~A Al~Obaidli, Salvatore Romeo,
  Kateryna Tymoshenko, and Antonio Uva. 2016.
\newblock Convkn at semeval-2016 task 3: Answer and question selection for
  question answering on arabic and english fora.
\newblock {\em SemEval\/} pages 896--903.

\bibitem[{Karpathy and Fei-Fei(2015)}]{karpathy2015deep}
Andrej Karpathy and Li~Fei-Fei. 2015.
\newblock Deep visual-semantic alignments for generating image descriptions.
\newblock In {\em CVPR\/}.

\bibitem[{Kumar et~al.(2016)Kumar, Irsoy, Su, Bradbury, English, Pierce,
  Ondruska, Gulrajani, and Socher}]{kumar2015ask}
Ankit Kumar, Ozan Irsoy, Jonathan Su, James Bradbury, Robert English, Brian
  Pierce, Peter Ondruska, Ishaan Gulrajani, and Richard Socher. 2016.
\newblock Ask me anything: Dynamic memory networks for natural language
  processing.
\newblock In {\em ICML\/}.

\bibitem[{Lai and Hockenmaier(2014)}]{lai2014illinois}
Alice Lai and Julia Hockenmaier. 2014.
\newblock Illinois-lh: A denotational and distributional approach to semantics.
\newblock {\em SemEval\/} .

\bibitem[{Marelli et~al.(2014)Marelli, Menini, Baroni, Bentivogli, Bernardi,
  and Zamparelli}]{marelli2014sick}
Marco Marelli, Stefano Menini, Marco Baroni, Luisa Bentivogli, Raffaella
  Bernardi, and Roberto Zamparelli. 2014.
\newblock A sick cure for the evaluation of compositional distributional
  semantic models.
\newblock In {\em LREC\/}.

\bibitem[{McClosky et~al.(2010)McClosky, Charniak, and
  Johnson}]{mcclosky2010automatic}
David McClosky, Eugene Charniak, and Mark Johnson. 2010.
\newblock Automatic domain adaptation for parsing.
\newblock In {\em NAACL-HLT\/}.

\bibitem[{Mihaylov and Nakov(2016)}]{mihaylov2016semanticz}
Todor Mihaylov and Preslav Nakov. 2016.
\newblock Semanticz at semeval-2016 task 3: Ranking relevant answers in
  community question answering using semantic similarity based on fine-tuned
  word embeddings.
\newblock {\em SemEval\/} pages 879--886.

\bibitem[{Mikolov et~al.(2013{\natexlab{a}})Mikolov, Chen, Corrado, and
  Dean}]{mikolov2013efficient}
Tomas Mikolov, Kai Chen, Greg Corrado, and Jeffrey Dean. 2013{\natexlab{a}}.
\newblock Efficient estimation of word representations in vector space.
\newblock In {\em ICLR\/}.

\bibitem[{Mikolov et~al.(2013{\natexlab{b}})Mikolov, Sutskever, Chen, Corrado,
  and Dean}]{mikolov2013distributed}
Tomas Mikolov, Ilya Sutskever, Kai Chen, Greg~S Corrado, and Jeff Dean.
  2013{\natexlab{b}}.
\newblock Distributed representations of words and phrases and their
  compositionality.
\newblock In {\em NIPS\/}.

\bibitem[{Miller et~al.(2016)Miller, Fisch, Dodge, Karimi, Bordes, and
  Weston}]{miller2016key}
Alexander Miller, Adam Fisch, Jesse Dodge, Amir-Hossein Karimi, Antoine Bordes,
  and Jason Weston. 2016.
\newblock Key-value memory networks for directly reading documents.
\newblock In {\em EMNLP\/}.

\bibitem[{Mou et~al.(2016)Mou, Meng, Yan, Li, Xu, Zhang, and
  Jin}]{mou2016transferable}
Lili Mou, Zhao Meng, Rui Yan, Ge~Li, Yan Xu, Lu~Zhang, and Zhi Jin. 2016.
\newblock How transferable are neural networks in nlp applications?
\newblock In {\em EMNLP\/}.

\bibitem[{Nakov et~al.(2016)Nakov, Màrquez, Moschitti, Magdy, Mubarak,
  Freihat, Glass, and Randeree}]{alessandromoschitti2016semeval}
Preslav Nakov, Lluís Màrquez, Alessandro Moschitti, Walid Magdy, Hamdy
  Mubarak, Abed~Alhakim Freihat, Jim Glass, and Bilal Randeree. 2016.
\newblock Semeval-2016 task 3: Community question answering.
\newblock {\em SemEval\/} pages 525--545.

\bibitem[{Nguyen et~al.(2016)Nguyen, Rosenberg, Song, Gao, Tiwary, Majumder,
  and Deng}]{nguyen2016ms}
Tri Nguyen, Mir Rosenberg, Xia Song, Jianfeng Gao, Saurabh Tiwary, Rangan
  Majumder, and Li~Deng. 2016.
\newblock Ms marco: A human generated machine reading comprehension dataset.
\newblock In {\em NIPS Workshop\/}.

\bibitem[{Pennington et~al.(2014)Pennington, Socher, and
  Manning}]{pennington2014glove}
Jeffrey Pennington, Richard Socher, and Christopher~D. Manning. 2014.
\newblock Glove: Global vectors for word representation.
\newblock In {\em EMNLP\/}.

\bibitem[{Rajpurkar et~al.(2016)Rajpurkar, Zhang, Lopyrev, and
  Liang}]{rajpurkar2016squad}
Pranav Rajpurkar, Jian Zhang, Konstantin Lopyrev, and Percy Liang. 2016.
\newblock Squad: 100,000+ questions for machine comprehension of text.
\newblock In {\em EMNLP\/}.

\bibitem[{Seo et~al.(2017)Seo, Kembhavi, Farhadi, and Hajishirzi}]{bidaf}
Minjoon Seo, Aniruddha Kembhavi, Ali Farhadi, and Hannaneh Hajishirzi. 2017.
\newblock Bidirectional attention flow for machine comprehension.
\newblock In {\em ICLR\/}.

\bibitem[{Trischler et~al.(2016)Trischler, Wang, Yuan, Harris, Sordoni,
  Bachman, and Suleman}]{trischler2016newsqa}
Adam Trischler, Tong Wang, Xingdi Yuan, Justin Harris, Alessandro Sordoni,
  Philip Bachman, and Kaheer Suleman. 2016.
\newblock Newsqa: A machine comprehension dataset.
\newblock {\em arXiv preprint arXiv:1611.09830\/} .

\bibitem[{Tymoshenko et~al.(2016)Tymoshenko, Bonadiman, and
  Moschitti}]{Tymoshenko2016Convolutional}
Kateryna Tymoshenko, Daniele Bonadiman, and Alessandro Moschitti. 2016.
\newblock Convolutional neural networks vs. convolution kernels: Feature
  engineering for answer sentence reranking.
\newblock In {\em NAACL-HLT\/}.

\bibitem[{Voorhees and Tice(2000)}]{qasent}
Ellen~M Voorhees and Dawn~M Tice. 2000.
\newblock Building a question answering test collection.
\newblock In {\em ACM SIGIR\/}.

\bibitem[{Wang and Jiang(2017{\natexlab{a}})}]{wang2016compare}
Shuohang Wang and Jing Jiang. 2017{\natexlab{a}}.
\newblock A compare-aggregate model for matching text sequences.
\newblock In {\em ICLR\/}.

\bibitem[{Wang and Jiang(2017{\natexlab{b}})}]{wang2016machine}
Shuohang Wang and Jing Jiang. 2017{\natexlab{b}}.
\newblock Machine comprehension using match-lstm and answer pointer.
\newblock In {\em ICLR\/}.

\bibitem[{Wang et~al.(2017)Wang, Hamza, and Florian}]{wang2017bilateral}
Zhiguo Wang, Wael Hamza, and Radu Florian. 2017.
\newblock Bilateral multi-perspective matching for natural language sentences.
\newblock {\em arXiv preprint arXiv:1702.03814\/} .

\bibitem[{Xiong et~al.(2017)Xiong, Zhong, and Socher}]{xiong2016dynamic}
Caiming Xiong, Victor Zhong, and Richard Socher. 2017.
\newblock Dynamic coattention networks for question answering.
\newblock In {\em ICLR\/}.

\bibitem[{Yang et~al.(2015)Yang, Yih, and Meek}]{yang2015wikiqa}
Yi~Yang, Wen-tau Yih, and Christopher Meek. 2015.
\newblock Wikiqa: A challenge dataset for open-domain question answering.
\newblock In {\em EMNLP\/}.

\bibitem[{Yin et~al.(2016)Yin, Sch{\"u}tze, Xiang, and Zhou}]{yin2015abcnn}
Wenpeng Yin, Hinrich Sch{\"u}tze, Bing Xiang, and Bowen Zhou. 2016.
\newblock Abcnn: Attention-based convolutional neural network for modeling
  sentence pairs.
\newblock {\em TACL\/} .

\bibitem[{Zeiler(2012)}]{zeiler2012adadelta}
Matthew~D Zeiler. 2012.
\newblock Adadelta: an adaptive learning rate method.
\newblock {\em arXiv preprint arXiv:1212.5701\/} .

\bibitem[{Zeiler and Fergus(2014)}]{zeiler2014visualizing}
Matthew~D Zeiler and Rob Fergus. 2014.
\newblock Visualizing and understanding convolutional networks.
\newblock In {\em ECCV\/}.

\bibitem[{Zhao et~al.(2014)Zhao, Zhu, and Lan}]{zhao2014ecnu}
Jiang Zhao, Tian~Tian Zhu, and Man Lan. 2014.
\newblock Ecnu: One stone two birds: Ensemble of heterogenous measures for
  semantic relatedness and textual entailment.
\newblock {\em SemEval\/} pages 271--277.

\end{thebibliography}
